%% file: root.tex
\documentclass[letterpaper, 10 pt, conference]{ieeeconf}  

\IEEEoverridecommandlockouts                              

\usepackage{graphics}
\usepackage{epsfig}
\usepackage{amsmath}
\usepackage{amssymb}
\usepackage{bm}
\usepackage{cite}
\usepackage{color}

\usepackage{algorithm}
\usepackage{algpseudocode}
\usepackage{booktabs}
\usepackage{tabularx}
\usepackage{array}

\usepackage{lipsum}

\usepackage[table]{xcolor}
\usepackage{pifont} 
\newcommand{\xmark}{\ding{55}}%
\DeclareMathAlphabet{\mathmybb}{U}{bbold}{m}{n}
\newcommand{\IndicatorFunction}{\mathmybb{1}}

\makeatletter
\let\NAT@parse\undefined
\makeatother

\usepackage{hyperref}
\usepackage{caption}

\DeclareCaptionFont{mysize}{\fontsize{8}{9.6}\selectfont}
\title{\LARGE \bf
DODGER: Safety-Guided Reinforcement Learning for Robot Navigation Among Dynamic Obstacles
}

\author{Sanghyuk Park$^{*1}$, Kwanwoo Lee$^{*1}$, Taekyung Kim$^{*2}$, Seohyeon Lim$^{3}$, and Yisoo Lee$^{\dagger4}$
\thanks{$^*$Equal contribution; $^\dagger$Corresponding author.}
\thanks{ This work was supported by the National Research Foundation of Korea (NRF) grant funded by the Korea government (MSIT) (RS-2026-25540546).}
\thanks{$^{1}$Sanghyuk Park and Kwanwoo Lee are with the Department of Intelligence and Information, Seoul National University, Republic of Korea. 
{\tt\footnotesize  sang0823, kwlee365@snu.ac.kr}}%
\thanks{$^{2}$Taekyung Kim is with the Department of Robotics, University of Michigan, Ann Arbor, MI, USA.
{\tt\footnotesize  taekyung@umich.edu}}%
\thanks{$^{3}$Seohyeon Lim is with the Department of Mechanical Engineering, Yonsei University, Republic of Korea. 
{\tt\footnotesize  joan0219@yonsei.ac.kr}}%
\thanks{$^{4}$Yisoo Lee is with the Center for Humanoid Research, Korea Institute of Science and Technology (KIST), Seoul, Republic of Korea.
{\tt\footnotesize  yisoo.lee@kist.re.kr}}%
}

\begin{document}

\maketitle
\thispagestyle{empty}
\pagestyle{empty}

\begin{abstract}
Robots operating in human-centered environments must safely navigate among multiple dynamic obstacles to avoid collisions with people and surrounding infrastructure. Control barrier functions (CBFs) provide an effective mechanism for safety filtering, and recent CBF-based reinforcement learning (RL) methods embed such safety information into learned policies. However, executing only safety-filtered actions during training can restrict policy exploration, a limitation that becomes particularly consequential in dynamic scenes where safety depends on relative robot-obstacle motion. We propose \textbf{DODGER}, a safety-guided RL framework that directly executes policy-generated actions to drive training rollouts while using CBF-filtered references and constraint violations to shape the policy toward collision-avoidance behavior. We evaluate DODGER through a Dubins-car safety analysis and demonstrate goal-directed navigation among multiple dynamic obstacles in full-order humanoid simulation and real-world humanoid experiments using LiDAR-based perception, without a runtime safety filter.
\href{https://psh0823.github.io/dodger-homepage}{\textcolor{red}{[Project Page]}}\footnote{Project page: \href{https://psh0823.github.io/dodger-homepage}{https://psh0823.github.io/dodger-homepage}} 
\href{https://github.com/psh0823/dodger}{\textcolor{red}{[Code]}}\footnote{Code: \href{https://github.com/psh0823/dodger}{https://github.com/psh0823/dodger}}  \href{https://dodger.taekyung.me}{\textcolor{red}{[Web Demo]}}

\end{abstract}

\section{INTRODUCTION}
\input{1_Introduction/intro}

\section{PRELIMINARIES}
\label{sec:preliminary}
\subsection{Control Barrier Functions}
\label{subsec:cbf}\input{2_Preliminary/a_cbf}
\subsection{CBF-RL}
\label{subsection:cbf_rl}\input{2_Preliminary/b_cbf_rl}

\begin{figure*}[!t]
    \centering
    \includegraphics[width=\textwidth]{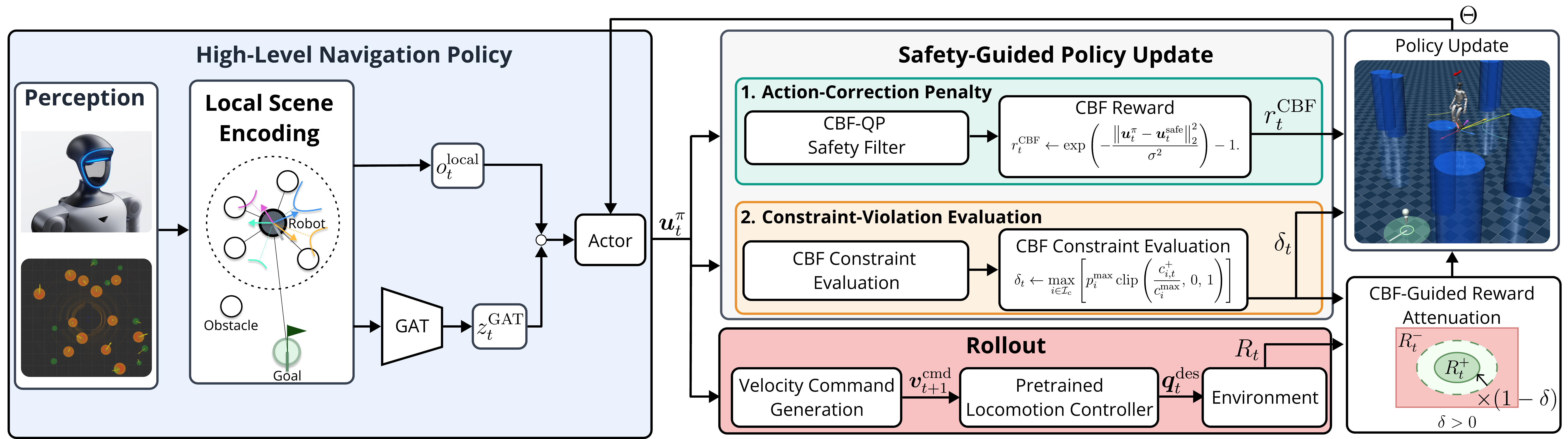}
    \caption{Overview of DODGER. The high-level policy combines local robot-goal
    information with a graph-based representation of surrounding obstacles to
    generate navigation commands. During training, the policy action directly drives the
    environment rollout, while the CBF-filtered reference and constraint violations guide learning. The resulting navigation command is executed
    by a fixed pretrained locomotion controller.}
    \label{fig:framework}
\end{figure*}

\section{METHODOLOGY}
\label{sec:method}
\input{3_Method/0_intro}
\subsection{Dynamic Parabolic CBF}
\label{subsec:dpcbf}\input{3_Method/a_dpcbf}
\subsection{DODGER Overview}
\label{subsec:dodger_overview}\input{3_Method/b_dodger_overview}
\subsection{Safety-Guided Reinforcement Learning}
\label{subsec:dodger_learning}\input{3_Method/c_dodger_learning}

\section{RESULTS}
\label{sec:results}
\subsection{Low-Dimensional Validation}
\label{subsec:low_dimensional_validation}\input{4_Results/a_dubins}
\subsection{Humanoid Dynamic-Obstacle Avoidance}
\label{subsec:humanoid_pedestrian_avoidance}\input{4_Results/b_simulation_comparison}

\begin{figure*}[!t]
    \centering
    \includegraphics[width=\textwidth]{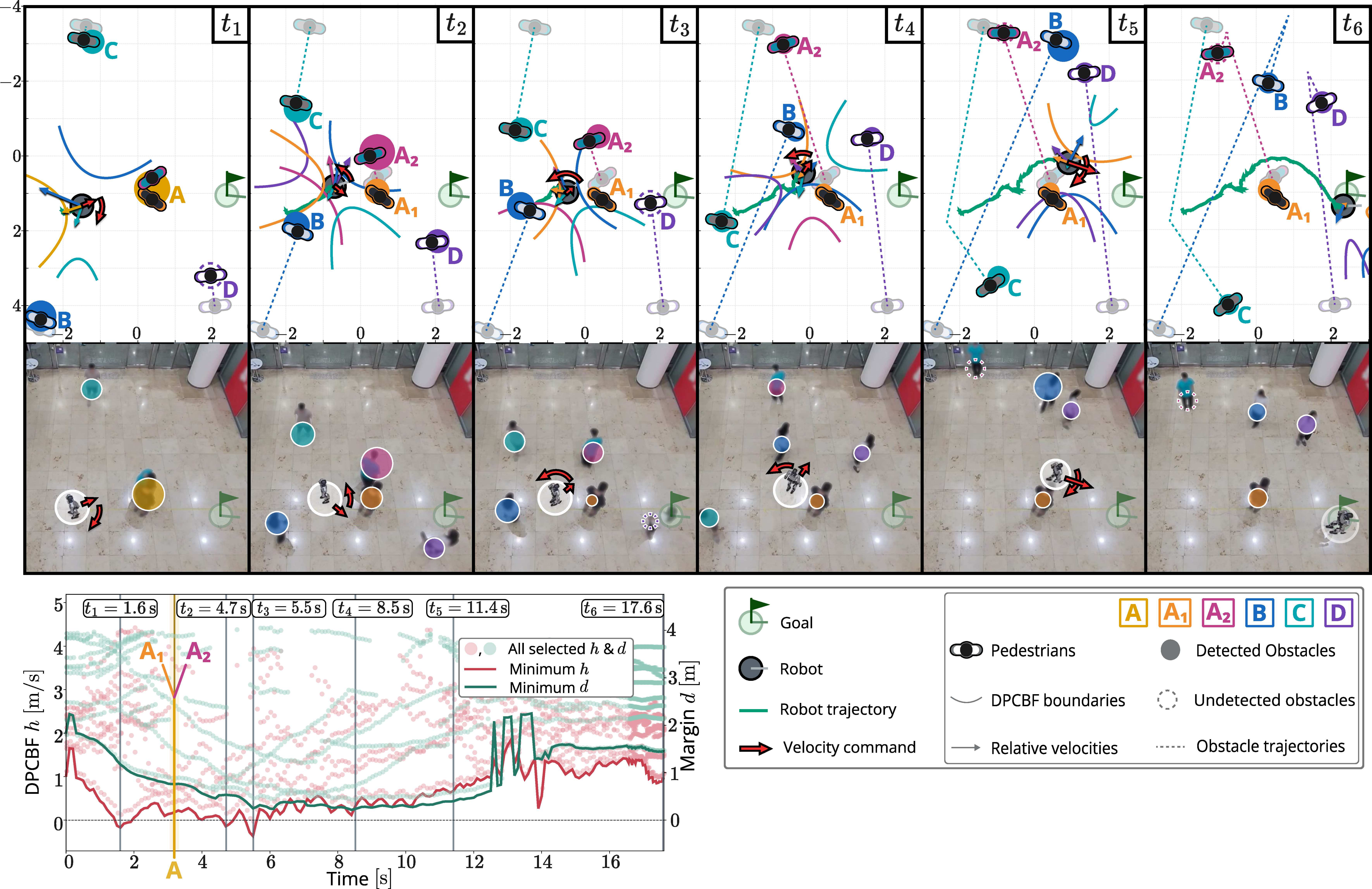}
    \caption{Real-robot navigation with a Unitree G1 among multiple moving
    pedestrians. The upper panels show the estimated scene and camera views at six
    representative times, together with DPCBF boundaries and relative velocities;
    $h<0$ when a relative velocity enters the unsafe side of its corresponding
    parabolic boundary. The lower-left plot shows the DPCBF values $h$ and margins
    $d$ of the selected obstacles, with solid curves denoting their minima.
    Although the minimum $h$ temporarily falls below zero, the minimum $d$
    remains positive throughout the experiment.}
    \label{fig:real_robot}
\end{figure*}

\subsection{Discussion}\input{4_Results/c_discussion}
\subsection{Real-Robot Experiments}\input{4_Results/d_real_robot_experiments}

\section{CONCLUSION}
\label{sec:conclusion}\input{5_Conclusion/conclusion}

\addtolength{\textheight}{0cm}

\bibliographystyle{IEEEtran}
\bibliography{references}

\clearpage
\onecolumn
\appendix
\subsection{Barrier Function Derivatives and Safety Conditions}
\label{app:lie_derivatives}\input{6_Appendix/a_lie_derivatives}
\subsection{Graph Encoding and Network Architecture}
\label{app:network_architecture}\input{6_Appendix/b_network_architecture}

\end{document}

%% file: 1_Introduction/intro.tex
Learning-based control has enabled agile and adaptive behaviors across diverse
robotic platforms, from aerial vehicles to legged robots~\cite{kaufmann2023champion,radosavovic2024real}. As robots increasingly operate in dynamic, human-centered environments, they must navigate toward task goals while safely interacting with moving obstacles. This requirement is particularly relevant to humanoid robots operating in human environments, where navigation often involves close interactions with moving people. For humanoids, dynamic-obstacle avoidance is typically handled through a low-dimensional navigation interface, while a separate whole-body controller realizes the commanded motion. The high-level navigation policy must therefore select goal-directed commands while accounting for the relative motion of surrounding obstacles.

Dynamic-obstacle avoidance is commonly addressed through online planning or safety filtering. MPC optimizes over a finite horizon using robot dynamics and predicted obstacle motion, but its computational cost increases with the planning horizon and the number of obstacles~\cite{andersson2016model,gaertner2021collisionfree}. CBF-based safety filters instead modify nominal commands online to satisfy prescribed safety constraints with minimal intervention~\cite{harms2025safe}. For hierarchical humanoid control, however, such constraints are typically formulated on a reduced-order navigation model, while the commanded motion is realized by the full-body system. Model mismatch and tracking error can therefore create a gap between the safety condition imposed on the high-level command and the realized robot motion~\cite{yang2025shield}. 
This motivates learning a navigation policy that is itself informed by the safety structure, rather than relying on online correction alone.

In Reinforcement Learning (RL), safety information is commonly incorporated through reward shaping or safety-filtered interaction~\cite{kim_your_2026}. Reward-based methods penalize constraint violations~\cite{nilaksh2024barrier}, but their effectiveness can depend strongly on the design and weighting of the penalty terms~\cite{wang2026reinforcement}. Safety-filtering methods instead modify policy actions before environment interaction to satisfy safety constraints~\cite{cheng2019end,emam2022safe,bejarano2025safety}. CBF-RL combines such filtering with a reward that penalizes deviations between nominal and filtered actions~\cite{yang2026cbfrlsafetyfilteringreinforcement}. However, because the filtered action is executed during training, the collected experience follows the state-action distribution induced by the safety filter rather than the policy itself. This can limit exposure to states arising from the policy's own actions, motivating a formulation that preserves policy-generated interaction while using CBF information only to guide learning.

\begin{figure}[!t]
    \centering
    \includegraphics[width=\linewidth]{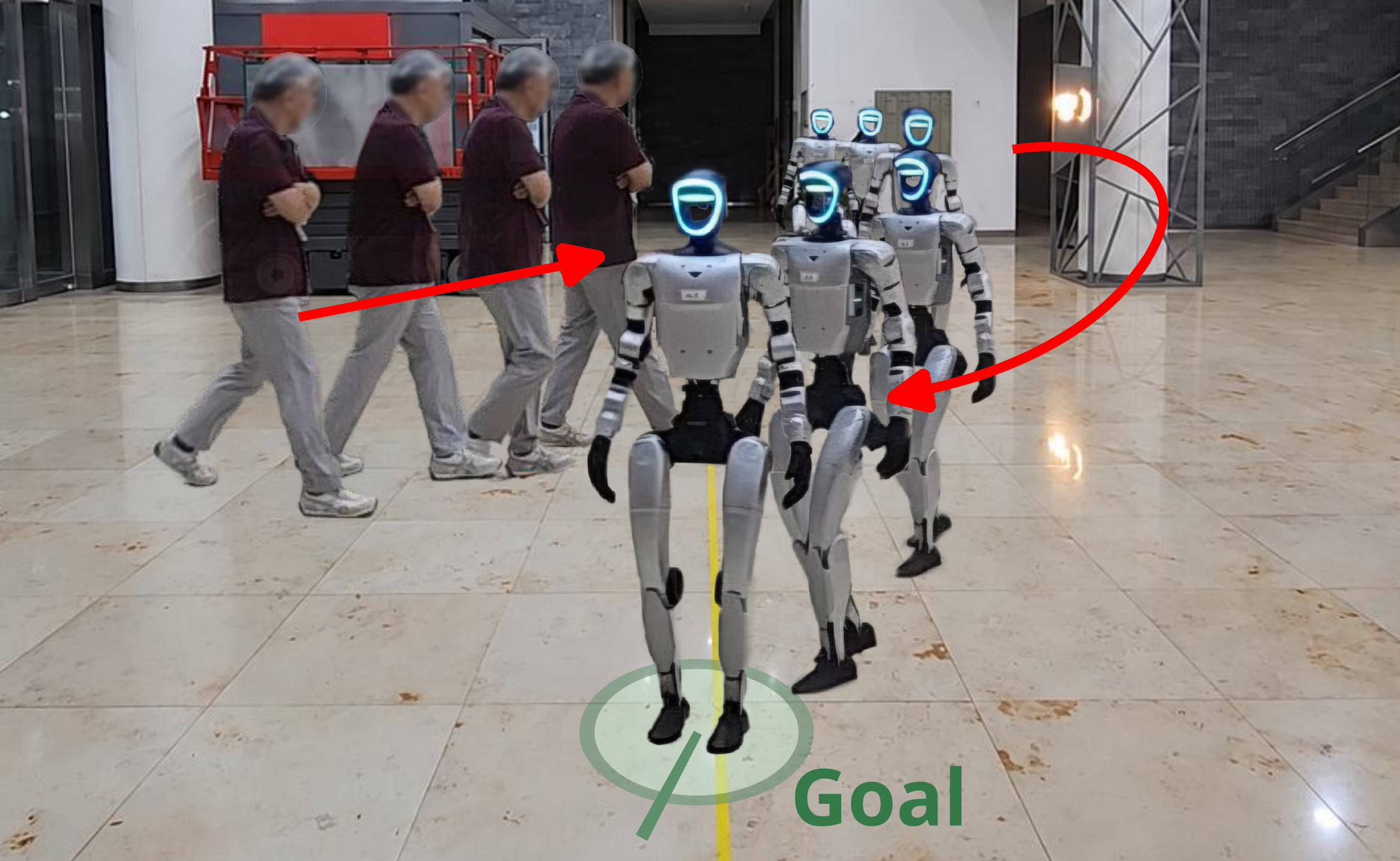}
    \caption{
    A humanoid avoids a moving pedestrian while navigating toward a goal using the DODGER policy. The policy is trained with CBF-based safety guidance and deployed without a runtime safety filter.}
    \label{fig:real_robot_1}
\end{figure}

To address these limitations, we propose \textbf{DODGER},
\textbf{D}ynamic \textbf{O}bstacle Avoidance with
\textbf{D}PCBF-\textbf{G}uided \textbf{E}xploration in
\textbf{R}einforcement Learning, a safety-guided RL framework for
goal-directed robot navigation among multiple dynamic obstacles. 
DODGER uses the Dynamic Parabolic CBF (DPCBF)~\cite{park2026collisionconesdynamicobstacle} to capture safety constraints induced by relative robot-obstacle motion, and integrates it with graph-based scene encoding and our safety-guided RL formulation.
Rather than executing safety-filtered actions during training,
DODGER directly rolls out policy-generated actions while using the filtered
reference and constraint violations to guide policy updates.

The main contributions of this work are:
\begin{itemize}
    \item We introduce a safety-guided RL training formulation in which policy-generated actions directly drive environment interaction, while CBF-filtered references and constraint evaluations are used only to guide policy learning. This allows the policy to learn from the consequences of its own actions without relying on safety-filtered rollouts.

    \item We introduce a constraint-aware return formulation that uses CBF and
    input-constraint violations to reduce positive returns, while a two-head
    critic preserves task-specific negative penalties.

    \item We develop DODGER for navigation among multiple dynamic obstacles using
    DPCBF-based safety guidance and graph-based scene encoding. We evaluate how
    the learned critic captures unsafe regions in a Dubins-car case study, and
    demonstrate goal-directed multi-obstacle navigation in full-order humanoid
    simulation and real-world deployment on a Unitree G1
    (Fig.~\ref{fig:real_robot_1}).
\end{itemize}

%% file: 2_Preliminary/a_cbf.tex
Consider a continuous-time control-affine system 
\begin{equation}
    \dot{\bm{x}}
    =
    f(\bm{x})
    +
    g(\bm{x})\bm{u},
    \label{eq:control_affine_system}
\end{equation}
where $\bm{x}\in\mathcal{X}\subseteq\mathbb{R}^{n}$ denotes the state and
$\bm{u}\in\mathcal{U}\subset\mathbb{R}^{m}$ denotes the control input.
The functions
$f:\mathcal{X}\rightarrow\mathbb{R}^{n}$ and
$g:\mathcal{X}\rightarrow\mathbb{R}^{n\times m}$
are assumed to be locally Lipschitz continuous.

Let $h: \mathcal{X} \rightarrow \mathbb{R}$ be a continuously differentiable function. The set $\mathcal{C}$ is defined as the zero superlevel set of $h$: $\mathcal{C}=\left\{\bm{x}\in\mathcal{X} \mid h(\bm{x})\geq0 \right\}$. A function $h$ is a control barrier function if there exists an extended class-$\mathcal{K}_{\infty}$ function $\alpha(\cdot):\mathbb{R}\rightarrow\mathbb{R}$ such that
\begin{equation}
    \sup_{\bm{u}\in\mathcal{U}}
    \left[
    L_f h(\bm{x})
        +
        L_g h(\bm{x})\bm{u}
        +
        \alpha\bigl(h(\bm{x})\bigr)
    \right]
    \geq 0,
    \quad
    \forall\bm{x}\in\mathcal{X},
    \nonumber
\end{equation}
where $L_fh$ and $L_gh$ denote the Lie derivatives of $h$ with respect to
$f$ and $g$, respectively. Under the standard regularity condition
$\nabla h(\bm{x})\neq0$ for all $\bm{x}\in\partial\mathcal{C}$, any locally
Lipschitz feedback controller $\bm{u}=\bm{k}(\bm{x})$ satisfying
$L_fh(\bm{x})+L_gh(\bm{x})\bm{k}(\bm{x})
+\alpha(h(\bm{x}))\geq0$
for all $\bm{x}\in\mathcal{X}$ renders $\mathcal{C}$ forward invariant for
the closed-loop system~\cite{8796030}. In practice, this constraint is
commonly enforced through a CBF-based Quadratic Program (CBF-QP) that minimally
modifies a nominal control input.

%% file: 2_Preliminary/b_cbf_rl.tex
CBF-RL~\cite{yang2026cbfrlsafetyfilteringreinforcement} integrates a
CBF-QP safety filter with RL during training and uses a
reward that encourages the actor output to remain close to the filtered
action. In its humanoid locomotion experiments, the CBF is constructed on a
reduced-order model while the policy is trained with the full-body dynamics.
The resulting policy is deployed without solving the CBF-QP online and
empirically exhibits fewer constraint violations.

However, two limitations remain. First, because CBF-RL executes the
safety-filtered action during training, the collected experience follows the
state-action distribution induced by the safety filter rather than that of the
unfiltered policy~\cite{yang2026cbfrlsafetyfilteringreinforcement}. This can
restrict exposure to states that would otherwise be encountered by the
unfiltered policy.
Second, extending this framework to dynamic-obstacle navigation requires a safety representation that explicitly accounts for the relative motion between the robot and surrounding obstacles. The reported CBF-RL experiments focus on static environments, whereas safety in dynamic scenes depends jointly on the relative position and velocity of surrounding obstacles.
These observations motivate preserving environment interaction under the policy's own actions
while incorporating a barrier formulation tailored to dynamic obstacle interactions.

%% file: 3_Method/0_intro.tex
To address these objectives, DODGER couples policy-generated environment
interaction with Dynamic Parabolic CBF (DPCBF)-derived safety guidance within a hierarchical
navigation framework.
The DPCBF is formulated on a reduced-order navigation model and used during training to provide a safety-filtered reference action and quantify constraint violations, while the actor action drives the environment rollout.
We first present the DPCBF formulation, followed by the overall DODGER architecture and the proposed learning framework.

%% file: 3_Method/a_dpcbf.tex
The DPCBF constructs a state-dependent parabolic safety boundary from the
relative position and velocity between the robot and surrounding obstacles.
Its velocity-dependent formulation provides a first-order safety constraint
for the reduced-order dynamics while adapting the boundary to dynamic
obstacle interactions.

\subsubsection{Reduced-Order Dynamics}
\label{subsubsec:dynamics_model}
We formulate navigation using a planar reduced-order model that captures
translational motion in the forward and lateral directions of a body-fixed
frame attached to the floating base, together with its heading motion. This
representation provides a low-dimensional interface for constructing the CBF
constraints independently of the full-order robot dynamics.

The state vector is given by
$\bm{x}=[x,y,\phi,v_f,v_l]^\top$, where $(x,y)$ denote the planar position
of the floating base, $\phi$ denotes its heading angle, and $v_f$ and $v_l$
denote the forward and lateral velocities expressed in the body frame,
respectively.

The reduced-order dynamics are given by
{\footnotesize
\begin{equation}
    \begin{bmatrix}
        \dot{x} \\ \dot{y} \\ \dot{\phi} \\ \dot{v}_f \\ \dot{v}_l
    \end{bmatrix}
    =
    \begin{bmatrix}
        v_f\cos\phi - v_l\sin\phi \\
        v_f\sin\phi + v_l\cos\phi \\
        0 \\
        0 \\
        0
    \end{bmatrix}
    +
    \begin{bmatrix}
        0 & 0 & 0 \\
        0 & 0 & 0 \\
        0 & 0 & 1 \\
        1 & 0 & 0 \\
        0 & 1 & 0
    \end{bmatrix}
    \begin{bmatrix}
        a_f \\ a_l \\ \omega
    \end{bmatrix},
    \label{eq:reduced_model_dynamics}
\end{equation}}
where the control input is
$\bm{u}=[a_f,a_l,\omega]^\top$.
Here, $a_f$ and $a_l$ denote the forward and lateral accelerations,
respectively, and $\omega$ denotes the yaw angular velocity.

\subsubsection{DPCBF Formulation}
\label{subsubsec:dpcbf_formulation}
We consider $N_{\mathrm{obs}}$ detected dynamic obstacles, with the $j$-th
obstacle state
\begin{equation}
    \bm{x}_{\mathrm{obs}}^{j}
    =
    \begin{bmatrix}
        x_{\mathrm{obs}}^{j} &
        y_{\mathrm{obs}}^{j} &
        \phi_{\mathrm{obs}}^{j} &
        v_{\mathrm{obs}}^{j}
    \end{bmatrix}^{\top},
    \label{eq:obstacle_state}
\end{equation}
where $(x_{\mathrm{obs}}^{j},y_{\mathrm{obs}}^{j})$, $\phi_{\mathrm{obs}}^{j}$,
and $v_{\mathrm{obs}}^{j}$ denote its center, heading, and forward speed,
respectively. Each obstacle is modeled with constant planar velocity during barrier
evaluation, i.e.,
$\dot{\phi}_{\mathrm{obs}}^{j}
=\dot{v}_{\mathrm{obs}}^{j}=0$~\cite{park2026collisionconesdynamicobstacle}. The superscript $j$ and explicit obstacle-state arguments are omitted
below. All Lie derivatives are evaluated along the joint robot--obstacle
dynamics.

Let
$\bm{p}_{\mathrm{rob}}=[x,y]^\top$ and
$\bm{p}_{\mathrm{obs}}=[x_{\mathrm{obs}},y_{\mathrm{obs}}]^\top$.
The relative position and velocity are
$\bm{p}_{\mathrm{rel}}=\bm{p}_{\mathrm{rob}}-\bm{p}_{\mathrm{obs}}$
and
$\bm{v}_{\mathrm{rel}}=\bm{v}_{\mathrm{rob}}-\bm{v}_{\mathrm{obs}}$,
where
$\bm{v}_{\mathrm{rob}}=[\dot{x},\dot{y}]^\top$
and
$\bm{v}_{\mathrm{obs}}
=v_{\mathrm{obs}}[\cos\phi_{\mathrm{obs}},
\sin\phi_{\mathrm{obs}}]^\top$.
Following DPCBF~\cite{park2026collisionconesdynamicobstacle},
$\bm{v}_{\mathrm{rel}}$ is expressed in a line-of-sight (LoS) frame whose
$\tilde{x}$-axis is aligned with $\bm{p}_{\mathrm{rel}}$, yielding
$\tilde{\bm{v}}_{\mathrm{rel}}
=
[\tilde{v}_{\mathrm{rel},x},
\tilde{v}_{\mathrm{rel},y}]^\top$.
We approximate the planar footprints of the robot and obstacle by circles
with radii $r_{\mathrm{rob}}$ and $r_{\mathrm{obs}}$, respectively, and
define the combined safety radius
$r=r_{\mathrm{rob}}+r_{\mathrm{obs}}$ and, for
$\lVert\bm{p}_{\mathrm{rel}}\rVert>r$, the distance-dependent margin
$d(\bm{x})=\sqrt{\lVert\bm{p}_{\mathrm{rel}}\rVert^2-r^2}$.
The DPCBF parameters are
$\lambda(\bm{x})
=k_{\lambda}d(\bm{x})/\lVert\bm{v}_{\mathrm{rel}}\rVert$
and
$\mu(\bm{x})=k_{\mu}d(\bm{x})$,
where $k_{\lambda},k_{\mu}>0$. In implementation,
$\lVert\bm{v}_{\mathrm{rel}}\rVert$ is lower-bounded by a small
$\epsilon_v>0$ to avoid numerical singularity. The resulting DPCBF is
\begin{equation}
    h(\bm{x},\bm{x}_{\mathrm{obs}})
    =
    \tilde{v}_{\mathrm{rel},x}
    +
    \lambda(\bm{x})\tilde{v}_{\mathrm{rel},y}^{2}
    +
    \mu(\bm{x}).
    \label{eq:dpcbf}
\end{equation}

The parameters $\lambda(\bm{x})$ and $\mu(\bm{x})$ adapt the parabolic
boundary to the relative robot--obstacle state: $\lambda(\bm{x})$ controls
its curvature, while $\mu(\bm{x})$ shifts its vertex along the
$\tilde{v}_{\mathrm{rel},x}$ axis. The boundary becomes less restrictive for
more distant obstacles or smaller relative speeds, and more conservative as
the obstacle approaches or the relative speed increases.

%% file: 3_Method/b_dodger_overview.tex
As illustrated in Fig.~\ref{fig:framework}, DODGER couples a high-level navigation policy with a pretrained low-level policy. The pretrained low-level policy is kept fixed during high-level DODGER training.
At each time step $t$, the high-level navigation policy receives a local
observation defined as
\begin{align}
\bm{o}_t^{\mathrm{local}}
=
\big[
    &(\bm{p}_{g,t}^{b})^{\top},
    \sin e_{\phi,t},
    \cos e_{\phi,t},
    (\bm{v}_{t}^{b})^{\top},
    \dot{\phi}_{t}, \notag\\
    &\hspace{3cm}
    (\bm{v}_{t}^{\mathrm{cmd}})^{\top},
    (\bm{u}_{t-1}^{\pi})^{\top}
\big]^{\top},
\label{eq:local_observation}
\end{align}
where superscript $b$ denotes body-frame coordinates,
$\bm{p}_{g,t}^{b}\in\mathbb{R}^{2}$ is the relative goal position, and
$e_{\phi,t}=\operatorname{wrap}(\phi_g-\phi_t)$ is the heading error
for goal heading $\phi_g$, wrapped to $[-\pi,\pi)$.
The measured planar velocity and yaw rate are
$\bm{v}_{t}^{b}=[v_{f,t},v_{l,t}]^\top$ and $\dot{\phi}_t$. The current velocity command is
$\bm{v}_{t}^{\mathrm{cmd}}
=
[v_{f,t}^{\mathrm{cmd}},
v_{l,t}^{\mathrm{cmd}},
\omega_{t}^{\mathrm{cmd}}]^{\top}$,
and $\bm{u}_{t-1}^{\pi}\in\mathbb{R}^{3}$ is the previous high-level actor
action.

To accommodate a varying number of surrounding obstacles, DODGER represents
the robot, goal, and detected obstacles as a graph
$\mathcal{G}_t=(\mathcal{V}_t,\mathcal{E}_t)$, with node set $\mathcal{V}_t$ and edge set $\mathcal{E}_t$. Following~\cite{kim2026learning},
we use robot-centered attention: the robot node serves as the query over the
goal and all detected obstacle nodes, with no message passing among non-robot
nodes. Each node corresponds to the robot, goal, or a detected obstacle, with
a one-hot encoding indicating its node type. The edge features encode the
relative position and velocity between connected nodes, together with the
minimum distance between their boundaries. A graph attention encoder (GAT)
aggregates these features into a robot-node embedding
\begin{equation}
    \bm{z}_t^{\mathrm{GAT}}
    =
    \Phi_{\psi}^{\mathrm{GAT}}(\mathcal{G}_t),
    \label{eq:gat_context}
\end{equation}
where $\Phi_{\psi}^{\mathrm{GAT}}$ denotes the graph attention encoder
parameterized by $\psi$. The high-level actor observation is then formed by
concatenating the local observation and graph embedding:
\begin{equation}
    \bm{o}_t
    =
    \begin{bmatrix}
        \bm{o}_t^{\mathrm{local}} \\
        \bm{z}_t^{\mathrm{GAT}}
    \end{bmatrix}.
    \label{eq:high_level_observation}
\end{equation}

During training, the high-level policy, parameterized by $\theta$, samples
\begin{equation}
    \bm{u}_t^\pi
    :=
    \begin{bmatrix}
        a_{f,t}^\pi & a_{l,t}^\pi & \omega_t^\pi
    \end{bmatrix}^{\top}
    \sim \pi_\theta(\cdot\mid\bm{o}_t),
    \label{eq:high_level_action}
\end{equation}
where $a_{f,t}^{\pi}$ and $a_{l,t}^{\pi}$ are integrated over the high-level
control interval $\Delta t>0$ to obtain the forward and lateral
velocity commands, while
$\omega_t^{\pi}$ is directly used as the yaw-rate command.
This action parameterization matches the reduced-order control input in
\eqref{eq:reduced_model_dynamics}, allowing DPCBF constraints to be evaluated
directly on the high-level actor action.
The resulting command
$\bm{v}_{t+1}^{\mathrm{cmd}}
:=
[v_{f,t}^{\mathrm{cmd}}+\Delta t\,a_{f,t}^{\pi},
v_{l,t}^{\mathrm{cmd}}+\Delta t\,a_{l,t}^{\pi},
\omega_{t}^{\pi}]^{\top}$
is passed, together with proprioceptive observations, to the low-level
policy, which generates the robot control commands $\bm{q}^{\text{des}}_t$.

%% file: 3_Method/c_dodger_learning.tex
Unlike CBF-RL~\cite{yang2026cbfrlsafetyfilteringreinforcement}, DODGER executes the actor's output directly during training and uses the CBF-filtered reference and constraint violations only to guide learning.

\subsubsection{Safety Filter}
\label{subsubsec:safety_filter}
During training, DODGER computes a QP reference that penalizes deviations
from the actor action $\bm{u}_t^{\pi}$ while incorporating DPCBF and input constraints.
Its unrelaxed formulation is
\begin{subequations}
\begin{align}
    \bm{u}_t^{\mathrm{safe}}
    &=
    \underset{\bm{u}_t}{\operatorname{argmin}}
    \quad
    \frac{1}{2}
    \left\|
        \bm{u}_t-\bm{u}_t^{\pi}
    \right\|_2^2
    \\
    \text{s.t.}\quad
    &
    L_f h_j(\bm{x}_t)
    +
    L_g h_j(\bm{x}_t)\bm{u}_t
    +
    \alpha\!\left(h_j(\bm{x}_t)\right)
    \geq 0,
    \notag\\
    &
    \hspace{3cm}
    \forall j\in\{1,\ldots,N_{\mathrm{obs}}\},
    \label{eq:safety_filter_cbf}
    \\
    &
    v_f^{\mathrm{lb}}
    \leq
    v_{f,t}^{\mathrm{cmd}}
    +
    \Delta t\,a_{f,t}
    \leq
    v_f^{\mathrm{ub}},
    \label{eq:safety_filter_vf}
    \\
    &
    v_l^{\mathrm{lb}}
    \leq
    v_{l,t}^{\mathrm{cmd}}
    +
    \Delta t\,a_{l,t}
    \leq
    v_l^{\mathrm{ub}},
    \label{eq:safety_filter_vl}
    \\
    &
    \omega^{\mathrm{lb}}
    \leq
    \omega_{t}
    \leq
    \omega^{\mathrm{ub}}.
    \label{eq:safety_filter_omega}
\end{align}
\end{subequations}
Here,
$\bm{u}_t=[a_{f,t},a_{l,t},\omega_t]^{\top}$
is the QP decision variable, and $h_j$ denotes the
DPCBF in \eqref{eq:dpcbf} constructed for the $j$-th detected obstacle.
The superscripts $(\cdot)^{\mathrm{lb}}$ and $(\cdot)^{\mathrm{ub}}$
denote the lower and upper command bounds. During training, the unrelaxed
QP was feasible in $99.6\%$ of solves; slack relaxation was used for the
remaining $0.4\%$.

\subsubsection{Safety-Guided RL Training}
\label{subsubsec:rl_training}
DODGER builds upon the Constraints as Terminations (CaT)
principle~\cite{10802334}, which maps soft constraint violations to
termination probabilities used in return estimation. Rather than executing
the CBF-filtered reference or terminating the rollout upon a constraint
violation, DODGER continues the rollout with the actor action
$\bm{u}^{\pi}_{t}$ and uses the resulting termination probability only when
computing the learning target.

Let $\mathcal{I}_c$ and $\mathcal{I}_r$ denote the index sets of soft
constraints and reward terms, respectively. We denote a constraint value by
$c_{i_c,t}=c_{i_c}(\bm{x}_t,\bm{u}_t^\pi)$, where positive values indicate
violations, and define
$c_{i_c,t}^{+}=\max(0,c_{i_c,t})$.

For the DPCBF constraint associated with the $j$-th obstacle, we define
\begin{equation}
    c^{\mathrm{DPCBF}}_{j,t}
    =
    -
    \left[
        L_f h_j(\bm{x}_t)
        +
        L_g h_j(\bm{x}_t)\bm{u}^{\pi}_{t}
        +
        \alpha\!\left(h_j(\bm{x}_t)\right)
    \right].
    \label{eq:dpcbf_violation}
\end{equation}
For the input constraints, let
\begin{equation}
    \bm{\xi}^{\pi}_{t}
    =
    \begin{bmatrix}
        v^{\mathrm{cmd}}_{f,t}+\Delta t\,a^{\pi}_{f,t}, &
        v^{\mathrm{cmd}}_{l,t}+\Delta t\,a^{\pi}_{l,t}, &
        \omega^{\pi}_{t}
    \end{bmatrix}^{\top},
\end{equation}
with corresponding lower and upper bounds $\bm{\xi}^{\mathrm{lb}}$ and $\bm{\xi}^{\mathrm{ub}}$.
The violation of the $k$-th input constraint is defined as
\begin{equation}
    c^{\mathrm{input}}_{k,t}
    =
    \max
    \left\{
        \xi^{\mathrm{lb}}_{k}-\xi^{\pi}_{k,t},
        \,
        \xi^{\pi}_{k,t}-\xi^{\mathrm{ub}}_{k}
    \right\}.
    \label{eq:input_violation}
\end{equation}

\begin{algorithm}[t] \footnotesize
\caption{DODGER Training Procedure}
\label{alg:training_procedure}
\begin{algorithmic}[1]

\State Initialize $\Theta$ and
$c_{i_c}^{\max}>0\;(i_c\in\mathcal{I}_c)$

\For{$\mathrm{step}=1$ to $N_{\mathrm{steps}}$}

    \State Initialize $\bm{v}^{\mathrm{cmd}}_0$, $\bm{o}_0$, and
    $\mathbf{B}\gets\emptyset$

    \For{$t=0$ to $T-1$}

        \State $\bm{u}_t^\pi\sim\pi_\theta(\cdot\mid\bm{o}_t)$

        \State $\displaystyle
        \bm{u}^{\mathrm{safe}}_t
        \gets
        \textsc{SafetyFilter}(\bm{u}^{\pi}_t)$

        \State $\displaystyle
        (\bm{o}_{t+1},d_t)
        \gets
        \textsc{EnvironmentUpdate}
        \left(
            \bm{o}_t,
            \bm{u}^{\pi}_t
        \right)$

        \ForAll{$i_c\in\mathcal{I}_c$}
            \State $c_{i_c,t}
            \gets
            c_{i_c}(\bm{x}_t,\bm{u}^{\pi}_t)$
            \State $\displaystyle
            c_{i_c,t}^{+}
            \gets
            \max
            \left\{
                0,\,
                c_{i_c,t}
            \right\}$
        \EndFor

        \State $\displaystyle
        \delta_t
        \gets
        \max_{i_c\in\mathcal{I}_c}
        \left[
            p_{i_c}^{\max}
            \operatorname{clip}
            \left(
                \frac{c_{i_c,t}^{+}}{c_{i_c}^{\max}},
                0,
                1
            \right)
        \right]$

        \State $\displaystyle
        r_t^{\mathrm{CBF}}
        \gets
        \exp
        \left(
            -\frac{
                \left\|
                    \bm{u}^{\pi}_t
                    -
                    \bm{u}^{\mathrm{safe}}_t
                \right\|_2^2
            }{\sigma^2}
        \right)
        -1$

        \State Compute
        $r_t^{i_r}\;(i_r\in\mathcal{I}_r)$,
        including $r_t^{\mathrm{CBF}}$

        \State $R_t^{+}\gets0,\quad R_t^{-}\gets0$

        \ForAll{$i_r\in\mathcal{I}_r$}
            \State
            $R_t^{+}
            \gets
            R_t^{+}+\max(r_t^{i_r},0)$
            \State
            $R_t^{-}
            \gets
            R_t^{-}+\min(r_t^{i_r},0)$
        \EndFor

        \State Store
        $\left(
            \bm{o}_t,
            \bm{u}^{\pi}_t,
            \delta_t,
            R_t^{+},
            R_t^{-},
            d_t,
            \bm{o}_{t+1}
        \right)$
        in $\mathbf{B}$

    \EndFor

    \ForAll{$i_c\in\mathcal{I}_c$}
        \State $\displaystyle
        c_{i_c}^{\max}
        \gets
        \tau_c c_{i_c}^{\max}
        +
        (1-\tau_c)
        \max_{0\leq t<T}c_{i_c,t}^{+}$
    \EndFor

    \State $\Theta
    \gets
    \textsc{UpdatePPO}(\Theta,\mathbf{B})$

\EndFor

\end{algorithmic}
\end{algorithm}

Following CaT, each constraint $c_{i_c,t}$ is associated with a maximum
termination probability $p_{i_c}^{\max}$ that controls its enforcement
strength. The termination probability is computed from the normalized
constraint violations as
\begin{equation}
    \delta_t
    =
    \max_{i_c\in\mathcal{I}_{c}}
    \left[
        p^{\max}_{i_c}
        \operatorname{clip}
        \left(
            \frac{c^{+}_{i_c,t}}{c^{\max}_{i_c}},
            0,1
        \right)
    \right].
    \label{eq:termination_probability}
\end{equation}
Larger violations increase $\delta_t$, reducing the positive return and
bootstrapping contributions while the rollout continues with the actor action
$\bm{u}_t^\pi$.

The normalization scale $c^{\max}_{i_c}$ is updated from the maximum constraint violation observed during each rollout using
\begin{equation}
    c^{\max}_{i_c}
    \leftarrow
    \tau_c c^{\max}_{i_c}
    +
    (1-\tau_c)
    \max_{0\leq t<T} c^{+}_{i_c,t},
    \label{eq:constraint_scale}
\end{equation}
where $\tau_c\in(0,1)$ is the exponential smoothing coefficient.

DODGER additionally uses the safety-filtered action as a reference through
\begin{equation}
    r^{\mathrm{CBF}}_t
    =
    \exp
    \left(
        -
        \frac{
        \left\|
            \bm{u}^{\pi}_{t}
            -
            \bm{u}^{\mathrm{safe}}_{t}
        \right\|_2^2
        }{\sigma^2}
    \right)
    -1,
    \label{eq:cbf_reward}
\end{equation}
which penalizes deviations of the actor action from the QP reference.

Directly applying the CaT factor $(1-\delta_t)$ to signed rewards can
undesirably weaken negative penalties, since multiplication by
$(1-\delta_t)$ reduces their magnitude. This is particularly relevant to navigation tasks, where a linear progress reward naturally becomes negative when the robot moves away from the goal, while terminal events such as collisions and workspace violations require penalties with different severities. Transforming all rewards into
positive quantities can instead distort their intended relative importance.

To prevent the CaT factor from weakening task-specific negative penalties,
DODGER separates the reward terms into positive and negative components and
employs a two-head critic. The positive and negative reward sums are denoted
by $R_t^{+}$ and $R_t^{-}$, respectively. Inspired by the multi-head value
architecture in~\cite{munn2026scalablemultiobjectiverobotreinforcement}, the
critic independently estimates $V^{+}$ and $V^{-}$. The factor
$(1-\delta_t)$ is applied only to the positive branch, while the negative
branch preserves the original penalty magnitudes.

The positive-branch temporal-difference (TD) error is
\begin{equation}
    \epsilon_t^{+}
    =
    (1-\delta_t)R_t^{+}+
    \gamma(1-d_t)(1-\delta_t)
    V^{+}(\bm{o}_{t+1})
    -
    V^{+}(\bm{o}_t), \nonumber 
\end{equation}
where $\gamma\in[0,1)$ is the discount factor and $d_t\in\{0,1\}$ indicates environment termination. The same factor $(1-\delta_t)$ is applied to the positive-branch return
estimation, whereas the negative branch follows the standard proximal policy optimization (PPO) update~\cite{schulman2017proximalpolicyoptimizationalgorithms}.
The critic loss is
$L_V=\frac{1}{2}L_V^{+}+\frac{1}{2}L_V^{-}$,
where $L_V^{+}$ and $L_V^{-}$ are the corresponding value losses. The two
advantages are combined and normalized for the PPO update as
$A_t=\operatorname{Normalize}(A_t^{+}+A_t^{-})$.

Thus, DODGER incorporates safety guidance through two complementary learning signals. The action-correction reward $r^{\mathrm{CBF}}_t$ penalizes deviations from the CBF-filtered reference and, because $r^{\mathrm{CBF}}_t\in(-1,0]$, contributes to the negative reward branch. In parallel, the constraint-dependent factor $\delta_t$ reduces the positive branch according to the severity of constraint violations. Neither mechanism replaces the actor action during environment
interaction, thereby preserving policy-generated rollouts. Algorithm~\ref{alg:training_procedure} summarizes training, with trainable
parameters $\Theta$, rollout buffer $\mathbf{B}$, rollout length $T$,
and $N_{\mathrm{steps}}$ policy updates.

%% file: 4_Results/a_dubins.tex
We first evaluate the safety representation learned by DODGER using a
low-dimensional Dubins-car benchmark following~\cite{seo2025uncertainty}.
Since classical grid-based HJ reachability scales exponentially with state
dimension, direct online application to our five-dimensional reduced-order
navigation model is computationally impractical~\cite{1463302}; the
three-dimensional benchmark therefore provides a tractable HJ reference.
The state is $\bm{x}=[x,y,\phi]^{\top}$ with dynamics
$\dot{\bm{x}}
=
[v\cos\phi,v\sin\phi,0]^{\top}
+
[0,0,1]^{\top}\omega$,
where $v=1\,\mathrm{m/s}$ and
$\omega\in[-1.25,1.25]\,\mathrm{rad/s}$.

The obstacle is a circle centered at the origin with radius
$r_{\mathrm{obs}}=0.5\,\mathrm{m}$, defining the occupied state region
$\mathcal{O}
=
\{\bm{x}\in\mathcal{X}\mid
x^2+y^2\leq r_{\mathrm{obs}}^2\}$.
Given the known dynamics, we compute a DPCBF-based HJ reference by
backward reachability from $\{h\le0\}$~\cite{1463302}, using
$V_{\mathrm{HJ}}(\bm{x},T)=h(\bm{x})$.
Here, $T$ is the reachability horizon, and
$V_{\mathrm{HJ}}(\bm{x})=V_{\mathrm{HJ}}(\bm{x},0)$.
We define the HJ safety boundary as
$\partial V_{\mathrm{HJ}}
=\{\bm{x}\in\mathcal{X}\mid V_{\mathrm{HJ}}(\bm{x})=0\}$.
Using $V_{\mathrm{HJ}}(\bm{x})\leq0$ for unsafe states,
the HJ reference unsafe set is
$
\mathcal{U}_{\mathrm{HJ}}
=
\{\bm{x}\in\mathcal{X}
\mid V_{\mathrm{HJ}}(\bm{x})\leq0\}.
$

For this diagnostic, both methods use negative-reward critics with
identical reward scaling, evaluated on the benchmark state $\bm{x}$.
Their critic-induced risk sets are
\[
    \hat{\mathcal{U}}_{\epsilon}
    =
    \mathcal{O}
    \cup
    \left\{
    \bm{x}\in\mathcal{X}\setminus\mathcal{O}
    \mid V^{-}(\bm{x})\leq\epsilon
    \right\},
\]
where the same threshold $\epsilon=-0.5$ is used for all methods.

We quantify its agreement with the HJ reference using
\[
    C_{\mathrm{unsafe}}
    =
    \frac{
    \mu(\hat{\mathcal{U}}_{\epsilon}\cap\mathcal{U}_{\mathrm{HJ}})
    }{
    \mu(\mathcal{U}_{\mathrm{HJ}})
    },
    \qquad
    C_{\mathrm{FP}}
    =
    \frac{
    \mu(\hat{\mathcal{U}}_{\epsilon}\setminus\mathcal{U}_{\mathrm{HJ}})
    }{
    \mu(\mathcal{U}_{\mathrm{HJ}})
    }.
\]
Here, $\mu$ denotes set measure, $C_{\mathrm{unsafe}}$ is unsafe-set
coverage, and $C_{\mathrm{FP}}$ is the false-positive measure normalized
by $\mu(\mathcal{U}_{\mathrm{HJ}})$.

\begin{figure}[t]
    \centering
    \includegraphics[width=\linewidth]{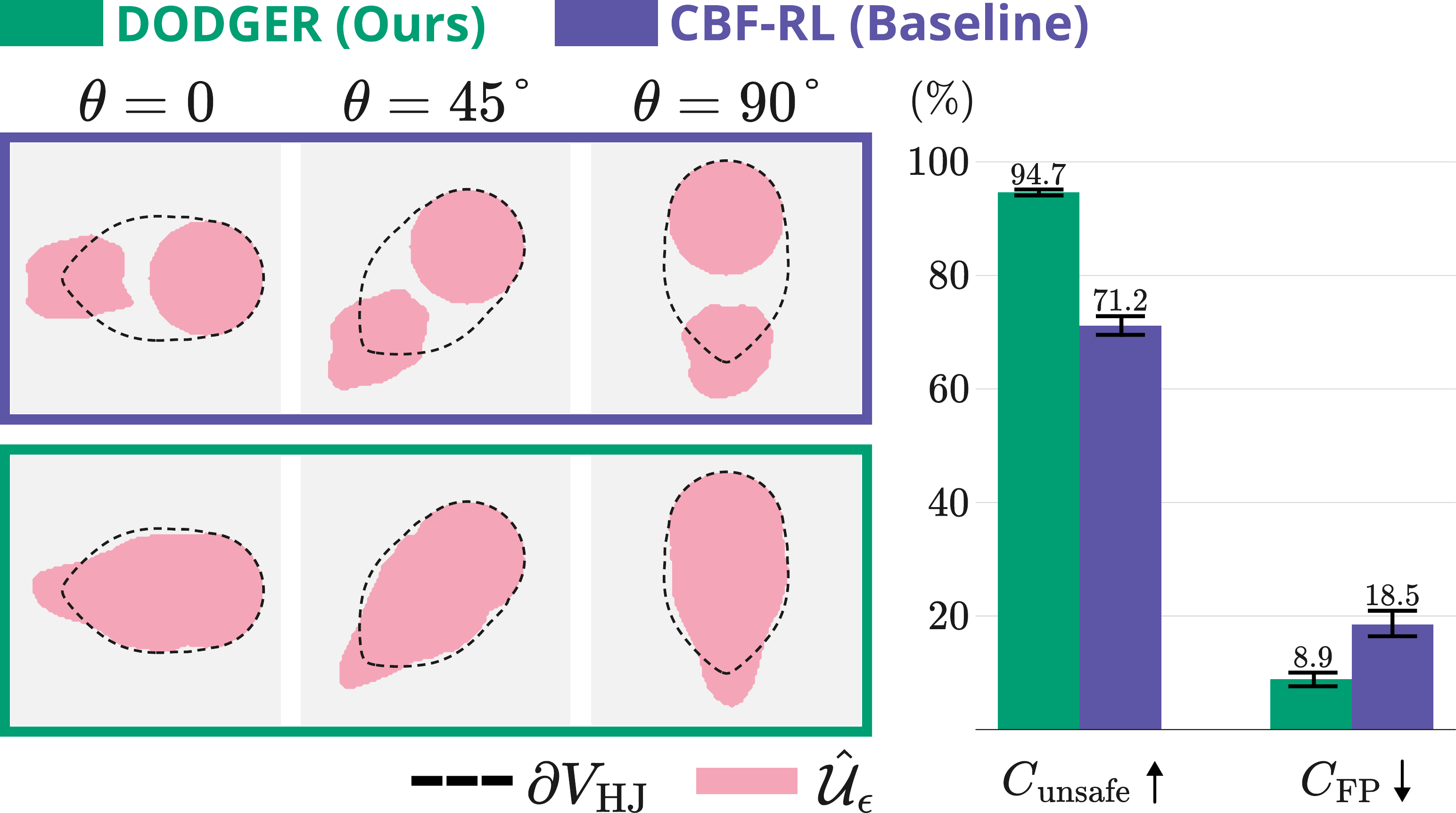}
    \caption{Comparison of DODGER and CBF-RL on the Dubins-car benchmark.
    The dashed black curve denotes the HJ safety boundary
    $\partial V_{\mathrm{HJ}}$, and the shaded region denotes
    $\hat{\mathcal{U}}_{\epsilon}$.}
    \label{fig:dubins_results}
\end{figure}

The reported metrics are averaged over five random seeds and are given as
mean $\pm$ standard deviation. As shown in Fig.~\ref{fig:dubins_results},
DODGER achieves
$C_{\mathrm{unsafe}}=94.7\pm0.93\%$ and
$C_{\mathrm{FP}}=8.9\pm2.47\%$, compared with
$71.2\pm3.47\%$ and $18.5\pm4.62\%$ for CBF-RL, respectively.
These results indicate that the critic-induced risk region learned by DODGER
more closely matches the HJ reference set, with higher coverage and fewer
false positives.

%% file: 4_Results/b_simulation_comparison.tex
\subsubsection{Experimental Setup}
We evaluate DODGER on full-order humanoid navigation among multiple dynamic
obstacles. The high-level navigation policy is trained in MuJoCo
Playground~\cite{zakka2025mujocoplayground} at $10\,\mathrm{Hz}$, while a
pretrained Unitree G1 velocity-tracking policy operates at $50\,\mathrm{Hz}$. We additionally provide an interactive web demo using MuJoCo compiled to
WebAssembly.\footnote{Web demo: \href{https://dodger.taekyung.me}{https://dodger.taekyung.me}}

Training uses a four-stage curriculum that progressively increases obstacle
density and velocity, state randomization, and observation noise, with the
final stage containing $20$ dynamic obstacles at speeds up to
$0.6\,\mathrm{m/s}$. At each high-level step, up to $10$ obstacles within
the $5\,\mathrm{m}$ sensing range are retained for GAT encoding and the
multi-constraint QP, prioritized by the alignment of relative velocity toward the robot
and by distance; unused
slots are padded and masked. The policy is trained for up to $600$ million
environment transitions and, after reaching the final curriculum stage,
evaluated every $3$ million transitions over $256$ episodes. An episode is
successful if the robot reaches the goal within $0.6\,\mathrm{m}$ before
failure or timeout while satisfying
$|\operatorname{wrap}(\phi_g-\phi_t)|\leq10^\circ$,
$\|\bm v_t^b\|_2\leq0.2\,\mathrm{m/s}$, and
$|\dot{\phi}_t|\leq0.2\,\mathrm{rad/s}$.
Convergence is defined as a mean success rate of at least $95\%$ over the
three most recent evaluations.

The command bounds are $v_f^{\mathrm{cmd}}\in[-1,2]\,\mathrm{m/s}$, $v_l^{\mathrm{cmd}}\in[-1,1]\,\mathrm{m/s}$, and $\omega^{\mathrm{cmd}}\in[-1,1]\,\mathrm{rad/s}$, while the actor accelerations are clipped to
$a_f^\pi\in[-4,4]\,\mathrm{m/s^2}$ and
$a_l^\pi\in[-2,2]\,\mathrm{m/s^2}$. For the CaT update, DPCBF and input-constraint violations are treated as soft
constraints, with $p_{i_c}^{\max}$ gradually increased to
$0.25$~\cite{10802334}.

At each of the five low-level steps, we recompute the QP reference and
constraint values. The CBF reward is averaged and the constraint violation
is max-pooled over the interval:
$r_t^{\mathrm{CBF}}
=
\frac{1}{5}\sum_{\ell=1}^{5}r_{t,\ell}^{\mathrm{CBF}}$
and
$c_{i_c,t}^{+}
=
\max_{\ell\in\{1,\ldots,5\}}c_{i_c,t,\ell}^{+}$.

The reward terms and their weights are summarized in Table~\ref{tab:reward_and_constraint_terms}.
Let $\bm{p}_t$ and $\bm{p}_g$ denote the world-frame robot and goal positions, and define
$\rho_t=\lVert\bm{p}_t-\bm{p}_g\rVert_2$,
$\Delta\rho_t=\rho_{t}-\rho_{t+1}$,
$\bar e_{\phi,t}=|\operatorname{wrap}(\phi_g-\phi_t)|$,
$\Delta\bar e_{\phi,t}
=\bar e_{\phi,t}-\bar e_{\phi,t+1}$,
$\Delta\bm{u}_t
=\bm{u}_t^\pi-\bm{u}_t^{\mathrm{safe}}$, and
$\Delta\widetilde{\bm{u}}_t^\pi
=\widetilde{\bm{u}}_t^\pi-\widetilde{\bm{u}}_{t-1}^\pi$. 
Here, $\widetilde{\bm{u}}_t^\pi$ denotes the componentwise normalized
actor action.
We use $\sigma=0.5$ and $\ell_h=1.0\,\mathrm{m}$.
Here, $\mathcal{A}_{\mathrm{hard}}$ denotes the prescribed workspace. Each
reward term is multiplied by its listed weight before being separated into
positive and negative components as described in
Sec.~\ref{subsubsec:rl_training}.

\begin{table}[t]
\centering
\caption{Reward terms used for training DODGER.}
\label{tab:reward_and_constraint_terms}
\footnotesize
\setlength{\tabcolsep}{3pt}
\renewcommand{\arraystretch}{1.1}
\begin{tabularx}{\columnwidth}{
    >{\centering\arraybackslash}p{0.20\columnwidth}
    >{\centering\arraybackslash}X
    >{\centering\arraybackslash}p{0.11\columnwidth}}
\toprule
\textbf{Term} & \textbf{Definition} & \textbf{Weight} \\
\midrule

$r^{\mathrm{CBF}}$
&
$\exp(-\lVert\Delta\bm{u}\rVert_2^2/\sigma^2)-1$
&
$100$
\\

$r^{\mathrm{prog}}$
&
$\Delta\rho$
&
$60$
\\

$r^{\mathrm{head}}$
&
$\Delta\bar e_{\phi}\exp[-(\rho/\ell_h)^2]$
&
$10$
\\

$r^{\mathrm{goal}}$
&
$\IndicatorFunction\{\text{goal reached}\}$
&
$20$
\\

$r^{\mathrm{obs}}$
&
$\IndicatorFunction\{\text{obstacle collision}\}$
&
$-55$
\\

$r^{\mathrm{outside}}$
&
$\IndicatorFunction\{\bm{p}\notin\mathcal{A}_{\mathrm{hard}}\}$
&
$-100$
\\

$r^{\mathrm{time}}$
&
$1$
&
$-0.01$
\\

$r^{\mathrm{timeout}}$
&
$\IndicatorFunction\{\text{timeout}\}$
&
$-15$
\\

$r^{\mathrm{rate}}$
&
$\lVert\Delta\widetilde{\bm{u}}^\pi\rVert_2^2$
&
$-0.02$
\\
\bottomrule
\end{tabularx}
\end{table}

Dynamic obstacles are detected and tracked within a $5\,\mathrm{m}$ radius
using LiDAR and the ROS~2 \texttt{obstacle\_detector\_2}
package~\cite{Przybya2017DetectionAT}.

\subsubsection{RL Training Comparison}
\label{subsubsec:rl_training_comparison}
We compare DODGER with two CBF-RL baselines under the same training configuration to evaluate learning efficiency in dense dynamic-obstacle navigation. Our CBF-RL baseline, adapted
from~\cite{yang2026cbfrlsafetyfilteringreinforcement}, executes
safety-filtered actions during training and sums the 
CBF-condition penalties over the detected obstacles.
Like CBF-RL, CBF-RL-Min executes the safety-filtered action during training,
but uses only the most critical obstacle-wise term, $\min_j\operatorname{clip}(\dot{h}_j+\alpha(h_j),-c,0)$,
with clipping bound $c>0$,
to avoid accumulating CBF penalties across multiple obstacles,
following~\cite{yang2026pacmanperceptionawarecbfrlwholebody}.

\begin{figure}[t]
    \centering
    \includegraphics[width=\linewidth]{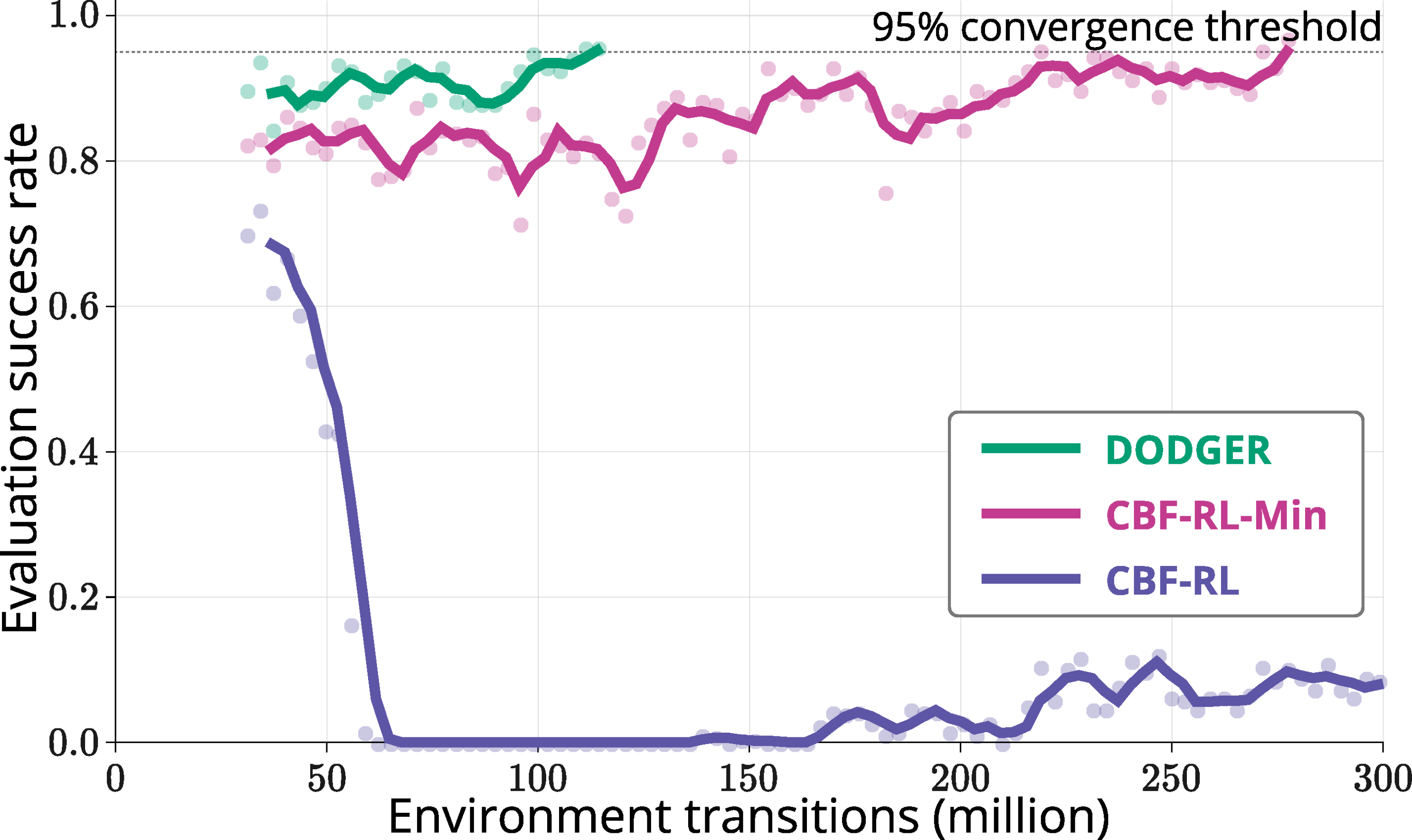}
    \caption{Evaluation success rates of DODGER, CBF-RL-Min, and CBF-RL
    during the first $300$ million environment transitions for humanoid
    navigation with obstacle speeds up to $0.6\,\mathrm{m/s}$. The markers denote individual evaluations,
    and the solid lines denote the moving average over the three most recent evaluations.} \label{fig:algorithm_comparison}
\end{figure}

As shown in Fig.~\ref{fig:algorithm_comparison}, DODGER converges after approximately $114$ million environment transitions, whereas CBF-RL-Min requires approximately $277$ million. In contrast, CBF-RL does not reach the convergence criterion within $600$ million transitions and achieves a final success rate of $72.3\%$ beyond the plotted range.

We further evaluate the methods at obstacle speeds up to
$0.8\,\mathrm{m/s}$ with a $1200$-million-transition budget. DODGER
converges after approximately $536$ million transitions, whereas CBF-RL-Min
and CBF-RL do not converge, with final success rates of $88.5\%$ and
$0.4\%$, respectively.

\subsubsection{CBF Formulation Comparison}
\label{subsubsec:cbf_comparison}
We compare DPCBF with the Collision Cone CBF
(C3BF)~\cite{11407993} and a distance-based CBF under the same DODGER
training configuration. C3BF incorporates relative position and velocity through
$h_{\mathrm{C3}}
=
\langle\bm{p}_{\mathrm{rel}},\bm{v}_{\mathrm{rel}}\rangle
+d(\bm{x})\lVert\bm{v}_{\mathrm{rel}}\rVert$.
In contrast, the distance-based barrier is defined as
$h_{\mathrm{dist}}=\lVert\bm{p}_{\mathrm{rel}}\rVert^2-r^2$.
Since $h_{\mathrm{dist}}$ has relative degree two with respect to the
reduced-order dynamics in \eqref{eq:reduced_model_dynamics}, we enforce the
distance-based safety constraint using an exponential CBF (ECBF)~\cite{7524935}.
We additionally evaluate a distance-based ECBF variant with yaw-rate regularization to examine the
large yaw commands observed with the original distance-based formulation.
All other components are kept identical across the CBF variants.

Table~\ref{tab:cbf_training_comparison} summarizes the training outcomes.
DPCBF and both distance-based ECBF variants reach curriculum stage~3 and
converge, with DPCBF requiring the fewest environment transitions. In
contrast, C3BF does not converge within the training budget and reaches only
stage~1.

\begin{table}[t]
\centering
\caption{Training outcomes under different CBF formulations. ``End'' denotes
the number of environment transitions (millions) at which training terminated;
curriculum stages are indexed from 0 to 3.}
\label{tab:cbf_training_comparison}
\footnotesize
\setlength{\tabcolsep}{3pt}
\renewcommand{\arraystretch}{1.15}
\begin{tabularx}{\columnwidth}{
    >{\raggedright\arraybackslash}X
    >{\centering\arraybackslash}p{0.17\columnwidth}
    >{\centering\arraybackslash}p{0.16\columnwidth}
    >{\centering\arraybackslash}p{0.12\columnwidth}}
\toprule
\textbf{CBF formulation} &
\textbf{End (M)} &
\textbf{Converged} &
\textbf{Stage} \\
\midrule
\rowcolor{gray!15}
DPCBF (DODGER)   & \textbf{113.97} & \checkmark & 3 \\
C3BF             & 600    & \xmark     & 1 \\
Dist.-ECBF       & 166.20 & \checkmark & 3 \\
Dist.-ECBF + yaw & 154.01 & \checkmark & 3 \\
\bottomrule
\end{tabularx}
\end{table}

To further characterize the yaw behavior, we compare the actor yaw commands
and the resulting G1 yaw rates for DPCBF and the two distance-based ECBF
variants. Over $N$ evaluated samples, the command root-mean-square
(RMS) value is
$\omega_{\mathrm{cmd,RMS}}
=
\sqrt{N^{-1}\sum_{t=1}^{N}(\omega_t^\pi)^2}$,
while the command near-saturation fraction is
$f_{\mathrm{cmd,sat}}
=
N^{-1}\sum_{t=1}^{N}
\IndicatorFunction\{|\omega_t^\pi|\geq0.9\,\mathrm{rad/s}\}$.
The G1 metrics use the measured yaw rate $\dot{\phi}_t$ and the same
threshold; fractions are reported as percentages.
As shown in Table~\ref{tab:cbf_yaw_comparison}, the distance-based ECBF variants produce larger yaw-rate commands and more frequent near-saturation than DPCBF, both in the actor output and in the resulting G1 motion.

\begin{table}[t]
\centering
\caption{Yaw-rate statistics for different CBF formulations.}
\label{tab:cbf_yaw_comparison}
\footnotesize
\setlength{\tabcolsep}{3pt}
\renewcommand{\arraystretch}{1.15}
\begin{tabularx}{\columnwidth}{
    >{\raggedright\arraybackslash}X
    >{\centering\arraybackslash}p{0.16\columnwidth}
    >{\centering\arraybackslash}p{0.18\columnwidth}
    >{\centering\arraybackslash}p{0.18\columnwidth}}
\toprule
\textbf{Metric} &
\cellcolor{gray!15}\textbf{DPCBF} &
\textbf{Dist.} &
\textbf{Dist.+Yaw} \\
\midrule
Cmd. RMS ($\mathrm{rad/s}$) & \cellcolor{gray!15} \textbf{0.554} & 0.934 & 0.894 \\
Cmd. near-sat. (\%)              & \cellcolor{gray!15} \textbf{5.48}  & 87.08 & 61.56 \\
G1 RMS ($\mathrm{rad/s}$)   & \cellcolor{gray!15} \textbf{0.435} & 0.808 & 0.763 \\
G1 near-sat. (\%)                & \cellcolor{gray!15} \textbf{0.87}  & 18.80 & 15.45 \\
\bottomrule
\end{tabularx}
\end{table}

%% file: 4_Results/c_discussion.tex
The graph-based encoder accommodates more than the training limit of
$10$ observed obstacles. Removing this cap at evaluation
yields a $97.27\%$ success rate, with $12.54$ observed obstacles on
average and up to $20$ simultaneously; $68.65\%$ of decisions involve
more than $10$ obstacles.

Across the low-dimensional HJ benchmark and full-order humanoid navigation,
the results support preserving unfiltered policy interaction: DODGER more
closely recovers the HJ reference region and converges faster than the
safety-filtered CBF-RL baselines.

The CBF comparison highlights the importance of the safety representation. C3BF achieves only a $23\%$ final success rate, with $76\%$ of episodes terminating by timeout and only $1\%$ by collision, indicating overly conservative behavior in dense scenes. The distance-based ECBF
converges but produces excessive rotational motion. For $h_{\mathrm{dist}}=\|\bm{p}_{\mathrm{rel}}\|^2-r^2$, the yaw coefficient in the higher-order CBF condition is
\begin{equation}
    \left.L_gL_fh_{\mathrm{dist}}\right|_{\omega}
    =
    2(p_{\mathrm{rel},y}\dot{x}-p_{\mathrm{rel},x}\dot{y}),
\end{equation}
which vanishes when the robot velocity is collinear with
$\bm{p}_{\mathrm{rel}}$, making the instantaneous ECBF constraint
insensitive to yaw in these configurations. Yaw regularization reduces this behavior but does not eliminate it. In contrast, DPCBF yields substantially lower yaw-rate RMS and near-saturation, suggesting that its LoS-based relative-motion representation provides more
effective directional guidance.

%% file: 4_Results/d_real_robot_experiments.tex
We deploy the DODGER policy trained with obstacle speeds up to
$0.8\,\mathrm{m/s}$ on a Unitree G1 among multiple moving pedestrians.
The floating-base pose is estimated from LiDAR and IMU measurements using
Direct LiDAR-Inertial Odometry (DLIO)~\cite{10160508}, and at most $10$
detected obstacles are selected per step for GAT encoding and DPCBF evaluation.

As shown in Fig.~\ref{fig:real_robot}, five pedestrians, denoted by
$A_1$, $A_2$, $B$, $C$, and $D$, move around the robot as it navigates
toward the goal. Real-world perception introduces merge--split errors and
abrupt changes in estimated obstacle geometry and velocity. For example,
$A_1$ and $A_2$ are initially perceived as a single obstacle $A$; although
they physically separate at approximately $3.18\,\mathrm{s}$, the detector
resolves them as $A_1$ and $A_2$ only later. Imperfect locomotion tracking
also causes deviations between commanded and realized robot motion.

The lower-left plot in Fig.~\ref{fig:real_robot} shows $h$ and $d$ for
the selected obstacles. The minimum $h$ is negative around $t_1$, $t_2$,
and $t_3$, whereas the minimum $d$ remains positive, indicating
separation of the estimated footprints despite temporary violations
of $h\ge0$. These negative values are associated with abrupt changes in perceived obstacle states, deviations from the constant-velocity obstacle model, and realized robot motion. Around $t_1$, pedestrians $A_1$ and $A_2$ are still represented as a single nearly stationary obstacle $A$; changes in its estimated radius induce a spurious nonzero velocity, causing $h<0$. Around $t_2$, pedestrian $B$ is estimated at $v_B=0.895\,\mathrm{m/s}$, exceeding the $0.8\,\mathrm{m/s}$ training range. After $A$ is split into $A_1$ and $A_2$, a similar geometry-induced velocity error occurs around $t_3$, where the nearly stationary $A_1$ is temporarily perceived as approaching the robot, producing the minimum observed DPCBF value while the distance-dependent margin remains positive ($d>0$). Despite these transient negative DPCBF values and perception errors, the robot avoids collision and reaches the goal.

%% file: 5_Conclusion/conclusion.tex
In this paper, we introduced DODGER, which combines direct policy interaction with DPCBF-based safety guidance to learn collision-avoidance behavior without requiring a runtime safety filter.
Low-dimensional validation showed close agreement with the HJ reference, while full-order humanoid simulations demonstrated efficient learning in dense multi-obstacle navigation and the effectiveness of DPCBF over alternative CBF formulations. Real-world humanoid experiments further showed successful navigation among multiple moving obstacles under perception and execution uncertainties.
Future work will extend the framework toward richer three-dimensional obstacle representations and full-order, perception-aware safety modeling for more reliable navigation in complex human environments.

%% file: 6_Appendix/a_lie_derivatives.tex
\subsubsection{Common Notation}
The obstacle index $j$ is omitted throughout this appendix for clarity.
Recall that
\[
    \bm{p}_{\mathrm{rel}}
    =
    \bm{p}_{\mathrm{rob}}
    -
    \bm{p}_{\mathrm{obs}},
    \qquad
    \bm{v}_{\mathrm{rel}}
    =
    \bm{v}_{\mathrm{rob}}
    -
    \bm{v}_{\mathrm{obs}},
\]
and
\[
    d(\bm{x})
    =
    \sqrt{
        \lVert\bm{p}_{\mathrm{rel}}\rVert^{2}
        -
        r^{2}
    }.
\]
Let
\[
    \beta
    =
    \operatorname{atan2}
    \left(
        p_{\mathrm{rel},y},
        p_{\mathrm{rel},x}
    \right)
\]
denote the line-of-sight (LoS) angle and define
$\tilde{\phi}=\phi-\beta$.
The relative velocity expressed in the LoS frame is
\begin{align}
    \tilde{v}_{\mathrm{rel},x}
    &=
    v_{\mathrm{rel},x}\cos\beta
    +
    v_{\mathrm{rel},y}\sin\beta,
    \\
    \tilde{v}_{\mathrm{rel},y}
    &=
    -v_{\mathrm{rel},x}\sin\beta
    +
    v_{\mathrm{rel},y}\cos\beta.
\end{align}
The DPCBF is
\begin{equation}
    h_{\mathrm{DPCBF}}
    =
    \tilde{v}_{\mathrm{rel},x}
    +
    k_{\lambda}
    \frac{d(\bm{x})}
    {\lVert\bm{v}_{\mathrm{rel}}\rVert}
    \tilde{v}_{\mathrm{rel},y}^{2}
    +
    k_{\mu}d(\bm{x}).
    \label{eq:app_dpcbf}
\end{equation}

\subsubsection{DPCBF for the Reduced-Order Model}
For the reduced-order dynamics in
\eqref{eq:reduced_model_dynamics}, the drift Lie derivative of
\eqref{eq:app_dpcbf} is
\begin{align}
    L_fh_{\mathrm{DPCBF}}
    &=
    \frac{
        \tilde{v}_{\mathrm{rel},y}^{2}
    }{
        \lVert\bm{p}_{\mathrm{rel}}\rVert
    }
    \nonumber\\
    &\quad+
    k_{\lambda}
    \frac{
        \tilde{v}_{\mathrm{rel},x}
        \tilde{v}_{\mathrm{rel},y}^{2}
    }{
        \lVert\bm{v}_{\mathrm{rel}}\rVert
    }
    \left(
        \frac{
            \lVert\bm{p}_{\mathrm{rel}}\rVert
        }{
            d(\bm{x})
        }
        -
        \frac{
            2d(\bm{x})
        }{
            \lVert\bm{p}_{\mathrm{rel}}\rVert
        }
    \right)
    \nonumber\\
    &\quad+
    k_{\mu}
    \frac{
        \lVert\bm{p}_{\mathrm{rel}}\rVert
    }{
        d(\bm{x})
    }
    \tilde{v}_{\mathrm{rel},x}.
    \label{eq:app_dpcbf_lf}
\end{align}

The input Lie derivative is written as
\begin{equation}
    L_gh_{\mathrm{DPCBF}}
    =
    \begin{bmatrix}
        \ell_f(\bm{x}) &
        \ell_l(\bm{x}) &
        \ell_{\omega}(\bm{x})
    \end{bmatrix},
    \label{eq:app_dpcbf_lg}
\end{equation}
where
\begin{align}
    \ell_f(\bm{x})
    &=
    \left(
        1
        -
        k_{\lambda}
        \frac{
            d(\bm{x})
        }{
            \lVert\bm{v}_{\mathrm{rel}}\rVert^{3}
        }
        \tilde{v}_{\mathrm{rel},x}
        \tilde{v}_{\mathrm{rel},y}^{2}
    \right)
    \cos\tilde{\phi}
    \nonumber\\
    &\quad+
    k_{\lambda}d(\bm{x})
    \left(
        \frac{
            2\tilde{v}_{\mathrm{rel},y}
        }{
            \lVert\bm{v}_{\mathrm{rel}}\rVert
        }
        -
        \frac{
            \tilde{v}_{\mathrm{rel},y}^{3}
        }{
            \lVert\bm{v}_{\mathrm{rel}}\rVert^{3}
        }
    \right)
    \sin\tilde{\phi},
    \label{eq:app_dpcbf_lgf}
    \\
    \ell_l(\bm{x})
    &=
    -
    \left(
        1
        -
        k_{\lambda}
        \frac{
            d(\bm{x})
        }{
            \lVert\bm{v}_{\mathrm{rel}}\rVert^{3}
        }
        \tilde{v}_{\mathrm{rel},x}
        \tilde{v}_{\mathrm{rel},y}^{2}
    \right)
    \sin\tilde{\phi}
    \nonumber\\
    &\quad+
    k_{\lambda}d(\bm{x})
    \left(
        \frac{
            2\tilde{v}_{\mathrm{rel},y}
        }{
            \lVert\bm{v}_{\mathrm{rel}}\rVert
        }
        -
        \frac{
            \tilde{v}_{\mathrm{rel},y}^{3}
        }{
            \lVert\bm{v}_{\mathrm{rel}}\rVert^{3}
        }
    \right)
    \cos\tilde{\phi},
    \label{eq:app_dpcbf_lgl}
    \\
    \ell_{\omega}(\bm{x})
    &=
    -
    \left(
        1
        -
        k_{\lambda}
        \frac{
            d(\bm{x})
        }{
            \lVert\bm{v}_{\mathrm{rel}}\rVert^{3}
        }
        \tilde{v}_{\mathrm{rel},x}
        \tilde{v}_{\mathrm{rel},y}^{2}
    \right)
    \left(
        v_f\sin\tilde{\phi}
        +
        v_l\cos\tilde{\phi}
    \right)
    \nonumber\\
    &\quad+
    k_{\lambda}d(\bm{x})
    \left(
        \frac{
            2\tilde{v}_{\mathrm{rel},y}
        }{
            \lVert\bm{v}_{\mathrm{rel}}\rVert
        }
        -
        \frac{
            \tilde{v}_{\mathrm{rel},y}^{3}
        }{
            \lVert\bm{v}_{\mathrm{rel}}\rVert^{3}
        }
    \right)
    \left(
        v_f\cos\tilde{\phi}
        -
        v_l\sin\tilde{\phi}
    \right).
    \label{eq:app_dpcbf_lgw}
\end{align}

Accordingly, the DPCBF condition used for the reduced-order model is
\begin{equation}
    L_fh_{\mathrm{DPCBF}}
    +
    L_gh_{\mathrm{DPCBF}}
    \begin{bmatrix}
        a_f & a_l & \omega
    \end{bmatrix}^{\top}
    +
    \alpha
    \left(
        h_{\mathrm{DPCBF}}
    \right)
    \geq 0.
    \label{eq:app_dpcbf_condition}
\end{equation}

\subsubsection{DPCBF for the Dubins-Car Model}
For the Dubins-car benchmark, the state and control are
$\bm{x}=[x,y,\phi]^{\top}$ and $\omega$, respectively, with constant
forward speed $v$. The obstacle is stationary, such that
\[
    \bm{v}_{\mathrm{rel}}
    =
    \begin{bmatrix}
        v\cos\phi &
        v\sin\phi
    \end{bmatrix}^{\top}.
\]
We denote the corresponding barrier by
$h_{\mathrm{DPCBF}}^{\mathrm{Dubins}}$.

Its drift Lie derivative is
\begin{align}
    L_f
    h_{\mathrm{DPCBF}}^{\mathrm{Dubins}}
    &=
    \frac{
        \tilde{v}_{\mathrm{rel},y}^{2}
    }{
        \lVert\bm{p}_{\mathrm{rel}}\rVert
    }
    \nonumber\\
    &\quad+
    k_{\lambda}
    \frac{
        \tilde{v}_{\mathrm{rel},x}
        \tilde{v}_{\mathrm{rel},y}^{2}
    }{
        \lVert\bm{v}_{\mathrm{rel}}\rVert
    }
    \left(
        \frac{
            \lVert\bm{p}_{\mathrm{rel}}\rVert
        }{
            d(\bm{x})
        }
        -
        \frac{
            2d(\bm{x})
        }{
            \lVert\bm{p}_{\mathrm{rel}}\rVert
        }
    \right)
    \nonumber\\
    &\quad+
    k_{\mu}
    \frac{
        \lVert\bm{p}_{\mathrm{rel}}\rVert
    }{
        d(\bm{x})
    }
    \tilde{v}_{\mathrm{rel},x},
    \label{eq:app_dubins_dpcbf_lf}
\end{align}
and its input Lie derivative is
\begin{equation}
    L_g
    h_{\mathrm{DPCBF}}^{\mathrm{Dubins}}
    =
    -
    \tilde{v}_{\mathrm{rel},y}
    +
    2k_{\lambda}
    \frac{
        d(\bm{x})
    }{
        \lVert\bm{v}_{\mathrm{rel}}\rVert
    }
    \tilde{v}_{\mathrm{rel},x}
    \tilde{v}_{\mathrm{rel},y}.
    \label{eq:app_dubins_dpcbf_lg}
\end{equation}

The corresponding first-order CBF condition is
\begin{equation}
    L_f
    h_{\mathrm{DPCBF}}^{\mathrm{Dubins}}
    +
    L_g
    h_{\mathrm{DPCBF}}^{\mathrm{Dubins}}
    \omega
    +
    \alpha
    \left(
        h_{\mathrm{DPCBF}}^{\mathrm{Dubins}}
    \right)
    \geq 0.
    \label{eq:app_dubins_dpcbf_condition}
\end{equation}

\subsubsection{Collision Cone CBF}
The Collision Cone CBF~(C3BF) used in the comparison is
\begin{equation}
    h_{\mathrm{C3}}
    =
    \bm{p}_{\mathrm{rel}}^{\top}
    \bm{v}_{\mathrm{rel}}
    +
    d(\bm{x})
    \lVert\bm{v}_{\mathrm{rel}}\rVert.
    \label{eq:app_c3bf}
\end{equation}
Its drift Lie derivative is
\begin{equation}
    L_fh_{\mathrm{C3}}
    =
    \lVert\bm{v}_{\mathrm{rel}}\rVert^{2}
    +
    \frac{
        \lVert\bm{v}_{\mathrm{rel}}\rVert
    }{
        d(\bm{x})
    }
    \bm{p}_{\mathrm{rel}}^{\top}
    \bm{v}_{\mathrm{rel}}.
    \label{eq:app_c3bf_lf}
\end{equation}

The input Lie derivative is
\begin{equation}
    L_gh_{\mathrm{C3}}
    =
    \begin{bmatrix}
        \ell_f^{\mathrm{C3}} &
        \ell_l^{\mathrm{C3}} &
        \ell_{\omega}^{\mathrm{C3}}
    \end{bmatrix},
    \label{eq:app_c3bf_lg}
\end{equation}
where
\begin{align}
    \ell_f^{\mathrm{C3}}
    &=
    \left(
        p_{\mathrm{rel},x}
        +
        \frac{
            d(\bm{x})
        }{
            \lVert\bm{v}_{\mathrm{rel}}\rVert
        }
        v_{\mathrm{rel},x}
    \right)\cos\phi
    \nonumber\\
    &\quad+
    \left(
        p_{\mathrm{rel},y}
        +
        \frac{
            d(\bm{x})
        }{
            \lVert\bm{v}_{\mathrm{rel}}\rVert
        }
        v_{\mathrm{rel},y}
    \right)\sin\phi,
    \label{eq:app_c3bf_lgf}
    \\
    \ell_l^{\mathrm{C3}}
    &=
    -
    \left(
        p_{\mathrm{rel},x}
        +
        \frac{
            d(\bm{x})
        }{
            \lVert\bm{v}_{\mathrm{rel}}\rVert
        }
        v_{\mathrm{rel},x}
    \right)\sin\phi
    \nonumber\\
    &\quad+
    \left(
        p_{\mathrm{rel},y}
        +
        \frac{
            d(\bm{x})
        }{
            \lVert\bm{v}_{\mathrm{rel}}\rVert
        }
        v_{\mathrm{rel},y}
    \right)\cos\phi,
    \label{eq:app_c3bf_lgl}
    \\
    \ell_{\omega}^{\mathrm{C3}}
    &=
    -
    \left(
        p_{\mathrm{rel},x}
        +
        \frac{
            d(\bm{x})
        }{
            \lVert\bm{v}_{\mathrm{rel}}\rVert
        }
        v_{\mathrm{rel},x}
    \right)\dot{y}
    \nonumber\\
    &\quad+
    \left(
        p_{\mathrm{rel},y}
        +
        \frac{
            d(\bm{x})
        }{
            \lVert\bm{v}_{\mathrm{rel}}\rVert
        }
        v_{\mathrm{rel},y}
    \right)\dot{x}.
    \label{eq:app_c3bf_lgw}
\end{align}

Thus,
\begin{equation}
    \dot{h}_{\mathrm{C3}}
    =
    L_fh_{\mathrm{C3}}
    +
    L_gh_{\mathrm{C3}}
    \begin{bmatrix}
        a_f & a_l & \omega
    \end{bmatrix}^{\top},
\end{equation}
and the corresponding CBF condition is
\begin{equation}
    L_fh_{\mathrm{C3}}
    +
    L_gh_{\mathrm{C3}}
    \begin{bmatrix}
        a_f & a_l & \omega
    \end{bmatrix}^{\top}
    +
    \alpha
    \left(
        h_{\mathrm{C3}}
    \right)
    \geq 0.
    \label{eq:app_c3bf_condition}
\end{equation}

\subsubsection{Distance-Based Exponential CBF}
The distance-based barrier is
\begin{equation}
    h_{\mathrm{dist}}
    =
    \lVert\bm{p}_{\mathrm{rel}}\rVert^{2}
    -
    r^{2}.
    \label{eq:app_dist_cbf}
\end{equation}
Its first-order Lie derivatives are
\begin{equation}
    L_fh_{\mathrm{dist}}
    =
    2
    \bm{p}_{\mathrm{rel}}^{\top}
    \bm{v}_{\mathrm{rel}},
    \qquad
    L_gh_{\mathrm{dist}}
    =
    \bm{0}.
    \label{eq:app_dist_lf}
\end{equation}
Therefore, the barrier has relative degree two with respect to the
reduced-order dynamics. The second-order Lie derivatives are
\begin{equation}
    L_f^{2}h_{\mathrm{dist}}
    =
    2
    \lVert\bm{v}_{\mathrm{rel}}\rVert^{2},
    \label{eq:app_dist_lf2}
\end{equation}
and
\begin{align}
    L_gL_fh_{\mathrm{dist}}
    =
    2
    \begin{bmatrix}
        p_{\mathrm{rel},x}\cos\phi
        +
        p_{\mathrm{rel},y}\sin\phi
        &
        -p_{\mathrm{rel},x}\sin\phi
        +
        p_{\mathrm{rel},y}\cos\phi
        &
        p_{\mathrm{rel},y}\dot{x}
        -
        p_{\mathrm{rel},x}\dot{y}
    \end{bmatrix}.
    \label{eq:app_dist_lglf}
\end{align}

Hence,
\begin{align}
    \ddot{h}_{\mathrm{dist}}
    &=
    L_f^{2}h_{\mathrm{dist}}
    +
    L_gL_fh_{\mathrm{dist}}
    \begin{bmatrix}
        a_f & a_l & \omega
    \end{bmatrix}^{\top}.
\end{align}
In particular, the coefficient associated with the yaw-rate input is
\begin{equation}
    \left.
    L_gL_fh_{\mathrm{dist}}
    \right|_{\omega}
    =
    2
    \left(
        p_{\mathrm{rel},y}\dot{x}
        -
        p_{\mathrm{rel},x}\dot{y}
    \right),
    \label{eq:app_dist_yaw}
\end{equation}
which is the expression used in the discussion of the yaw behavior.

The distance-based safety constraint is enforced using a second-order
exponential CBF~(ECBF):
\begin{equation}
    L_f^{2}h_{\mathrm{dist}}
    +
    L_gL_fh_{\mathrm{dist}}
    \begin{bmatrix}
        a_f & a_l & \omega
    \end{bmatrix}^{\top}
    +
    (\alpha_1+\alpha_2)
    L_fh_{\mathrm{dist}}
    +
    \alpha_1\alpha_2
    h_{\mathrm{dist}}
    \geq 0,
    \label{eq:app_dist_ecbf_condition}
\end{equation}
where $\alpha_1>0$ and $\alpha_2>0$ are the ECBF gains.

%% file: 6_Appendix/b_network_architecture.tex
The high-level actor consists of a robot-centered graph attention encoder
followed by an MLP policy head. During training, the input graph contains the
robot, the goal, and at most $10$ selected obstacles. Each node is represented
as
\begin{equation}
    \bm{n}_{j}
    =
    \left[
        \bm{\ell}_{j}^{\top},
        (\bm{p}_{j}^{b})^{\top},
        r_j,
        (\bm{v}_{j}^{b})^{\top},
        m_j
    \right]^{\top}
    \in\mathbb{R}^{9},
    \label{eq:app_gat_node_feature}
\end{equation}
where $\bm{\ell}_{j}\in\mathbb{R}^{3}$ is a one-hot encoding identifying the
robot, obstacle, or goal node;
$\bm{p}_{j}^{b}\in\mathbb{R}^{2}$ and
$\bm{v}_{j}^{b}\in\mathbb{R}^{2}$ denote the body-frame position and velocity,
respectively; $r_j$ is the node radius; and $m_j\in\{0,1\}$ is the validity
mask used for padded obstacle slots.

We employ robot-centered attention, in which only the robot-node embedding is
computed. For each obstacle or goal node $j$, the edge from the robot node
$r$ is represented as
\begin{equation}
    \bm{e}_{rj}
    =
    \left[
        \bm{\ell}_{r}^{\top},
        \bm{\ell}_{j}^{\top},
        (\bm{p}_{j}^{b}-\bm{p}_{r}^{b})^{\top},
        d_{rj},
        (\bm{v}_{j}^{b}-\bm{v}_{r}^{b})^{\top}
    \right]^{\top}
    \in\mathbb{R}^{11},
    \label{eq:app_gat_edge_feature}
\end{equation}
where
\begin{equation}
    d_{rj}
    =
    \left\|
        \bm{p}_{j}^{b}-\bm{p}_{r}^{b}
    \right\|_2
    -
    (r_r+r_j)
    \label{eq:app_gat_surface_distance}
\end{equation}
denotes the minimum distance between the node boundaries. Thus, the
$11$-dimensional edge feature consists of the two $3$-dimensional node-type
labels, the $2$-dimensional relative position, the scalar boundary distance,
and the $2$-dimensional relative velocity. The validity mask $m_j$ is not
passed through the edge MLP; it is used only to mask the attention logits of
invalid padded nodes.

The attention mechanism first constructs an intermediate message
$\bm{m}_{rj}=\psi_1(\bm{e}_{rj})$. The attention weight assigned to node $j$
is computed as
\begin{equation}
    \beta_{rj}
    =
    \frac{
        m_j\exp\!\left(\psi_2(\bm{m}_{rj})\right)
    }{
        \sum_{k}
        m_k\exp\!\left(\psi_2(\bm{m}_{rk})\right)
    },
    \label{eq:app_gat_attention}
\end{equation}
and the resulting robot-node embedding is
\begin{equation}
    \bm{z}^{\mathrm{GAT}}
    =
    \sum_j
    \beta_{rj}\psi_3(\bm{m}_{rj})
    \in\mathbb{R}^{16}.
    \label{eq:app_gat_embedding}
\end{equation}

The detailed actor, critic, and frozen low-level policy architectures are
summarized in Table~\ref{tab:network_architecture}. The actor concatenates the
$16$-dimensional graph embedding with the $13$-dimensional local observation.
The critic directly receives the corresponding noise-free privileged
observation and uses a shared backbone with separate positive- and
negative-value heads. The pretrained low-level locomotion policy remains
fixed throughout high-level policy training.

\begin{table}
\centering
\caption{Network architectures used for humanoid navigation.}
\label{tab:network_architecture}
\footnotesize
\setlength{\tabcolsep}{3pt}
\renewcommand{\arraystretch}{1.15}
\begin{tabularx}{\columnwidth}{
    >{\raggedright\arraybackslash}p{0.28\columnwidth}
    >{\raggedright\arraybackslash}X}
\toprule
\textbf{Component} & \textbf{Configuration} \\
\midrule

\multicolumn{2}{l}{\textit{Graph Attention Encoder}} \\

Graph input
&
Robot, goal, and up to $10$ obstacle nodes
\\

Node representation
&
$\bm{n}_j\in\mathbb{R}^{9}$
\\

Edge representation
&
$\bm{e}_{rj}\in\mathbb{R}^{11}$
\\

Message network $\psi_1$
&
Linear$(11\!\rightarrow\!64)$ + ReLU,
Linear$(64\!\rightarrow\!16)$
\\

Attention network $\psi_2$
&
Linear$(16\!\rightarrow\!16)$ + ReLU,
Linear$(16\!\rightarrow\!1)$,
masked softmax
\\

Encoding network $\psi_3$
&
Linear$(16\!\rightarrow\!64)$ + ReLU,
Linear$(64\!\rightarrow\!16)$
\\

GAT output
&
Attention-weighted robot embedding
$\bm{z}^{\mathrm{GAT}}\in\mathbb{R}^{16}$
\\

\midrule
\multicolumn{2}{l}{\textit{High-Level Actor}} \\

Actor input
&
Concat$\bigl(
\bm{z}^{\mathrm{GAT}}[16],
\bm{o}^{\mathrm{local}}[13]
\bigr)=29$
\\

Hidden layer 1
&
Linear$(29\!\rightarrow\!256)$ + ELU
\\

Hidden layer 2
&
Linear$(256\!\rightarrow\!128)$ + ELU
\\

Hidden layer 3
&
Linear$(128\!\rightarrow\!64)$ + ELU
\\

Output layer
&
Linear$(64\!\rightarrow\!6)$
\\

Policy distribution
&
Three-dimensional tanh-squashed normal distribution
\\

\midrule
\multicolumn{2}{l}{\textit{Two-Head Critic}} \\

Critic input
&
Privileged observation:
$12\times9+13=121$
\\

Shared hidden layer 1
&
Linear$(121\!\rightarrow\!256)$ + ELU
\\

Shared hidden layer 2
&
Linear$(256\!\rightarrow\!256)$ + ELU
\\

Shared hidden layer 3
&
Linear$(256\!\rightarrow\!128)$ + ELU
\\

Positive-value head
&
Linear$(128\!\rightarrow\!1)$
\\

Negative-value head
&
Linear$(128\!\rightarrow\!1)$
\\

\bottomrule
\end{tabularx}
\end{table}